\documentclass{article}
\usepackage{ijcai22}

\usepackage{times}
\usepackage{soul}
\usepackage{url}
\usepackage[hidelinks]{hyperref}
\usepackage[utf8]{inputenc}
\usepackage[small]{caption}
\usepackage{graphicx}
\usepackage{amsmath}
\usepackage{amssymb}
\usepackage{amsthm}
\usepackage{booktabs}
\usepackage{multirow}
\usepackage{subcaption}
\usepackage{algorithm}
\usepackage{algorithmic}
\usepackage{float}
\usepackage[T1]{fontenc}
\usepackage{xspace}
\makeatletter
\DeclareRobustCommand\onedot{\futurelet\@let@token\@onedot}
\def\@onedot{\ifx\@let@token.\else.\null\fi\xspace}
\newcommand*{\rom}[1]{\expandafter\@slowromancap\romannumeral #1@}
\def\eg{\emph{e.g}\onedot} 
\def\ie{\emph{i.e}\onedot}

\usepackage{mathtools,xparse}

\title{Multiple Emotion Descriptors Estimation at the ABAW3 Challenge}

\author{
Didan Deng$^1$\footnote{Contact Author}
\affiliations
$^1$Hong Kong University of Science and Technology\\
\emails
ddeng@connect.ust.hk
}

\begin{document}

\maketitle

\begin{abstract}
To describe complex emotional states, psychologists have proposed multiple emotion descriptors: sparse descriptors like facial action units; continuous descriptors like valence and arousal; and discrete class descriptors like happiness and anger. According to \cite{ekman1969repertoire}, facial action units are sign vehicles that convey the emotion message, while discrete or continuous emotion descriptors are the messages perceived and expressed by human.

In this paper, we designed an architecture for multiple emotion descriptors estimation in participating the ABAW3 Challenge. Based on the theory of \cite{ekman1969repertoire}, we designed distinct architectures to measure the sign vehicles (\ie, facial action units) and the messages (\ie, discrete emotions, valence and arousal) given their different properties. The quantitative experiments on the ABAW3 challenge~\cite{kollias2022abaw} dataset has shown the superior performance of our approach over two baseline models.

\end{abstract}

\section{Introduction}
\label{sec:intro}

Automatic emotion recognition techniques in computer vision are built to estimate the emotion descriptors used in psychological studies: facial expressions (\eg, happiness), continuous emotions (\eg, valence and arousal) and facial action units (activation of \eg nose wrinkles). \cite{ekman1969repertoire} divided them into two categories: sign vehicles and messages. When presented with a smiling face, the observer measuring the sign vehicles would code the face as having a upward movement of the lip corners, while the observer measuring the messages would describe the face as "happiness" or positive valence.

The main difference between measuring sign vehicles and measuring messages is the focus. Measuring sign vehicles focuses on the facial behavior itself, not on the perception of the face. Measuring the messages focuses on the person observing the face or the messages obtained, not on the face itself \cite{ekman1969repertoire}. Therefore, we can assume that the action units (AU) detection is less dependent on the observer, while the estimation of facial expressions (EXPR), valence, and arousal (VA) is more dependent on the observer. This means it is more likely that the estimations of EXPR and VA will vary across different observers.

As many psychologists argued~\cite{luck1998role,combs2004role}, our attention system allows us to focus on something specific while filtering out unimportant details. It also affects our perception of the stimuli. It is reasonable to assume that different observers have different attention preferences. Given a smiling but frowning face, some people may focus on the frown and perceive it as negative emotions, while other people may focus on the smile and perceive it as positive emotions. Such disagreement is common for the face input conveying contradictory or vague messages. While the observers' attention preferences are often unknown, we can infer the disagreement by checking the messages conveyed by different regions of the face input. Furthermore, we can obtain the consensus of emotional messages by merging the messages conveyed by different regions together. Therefore, we propose to project the visual features of different facial regions to some shared feature space and average them. The shared feature space, which we call as the message space, is the metric space for facial expression and valence/arousal. 

For sign vehicles (action units) detection, each AU has its own relevant facial region. For example, the AU12 (lip corner puller) is closely related to the region around the mouth. In the meanwhile, there is correlation among some AUs because of their co-occurrence (or the opposite) in expressing emotions. For the sign vehicles space learning, we adapted the AU detection network proposed by~\cite{jacob2021facial}, as it learns both intra-AU attention and the inter-AU attention. 

Our main goal is to employ the psychological knowledge about multiple emotion descriptors into the architecture design in order to regularize the multiple emotion descriptors estimation, especially in the lack of complete emotion annotations.  
\begin{figure*}
    \centering
    \includegraphics[width=0.8\linewidth]{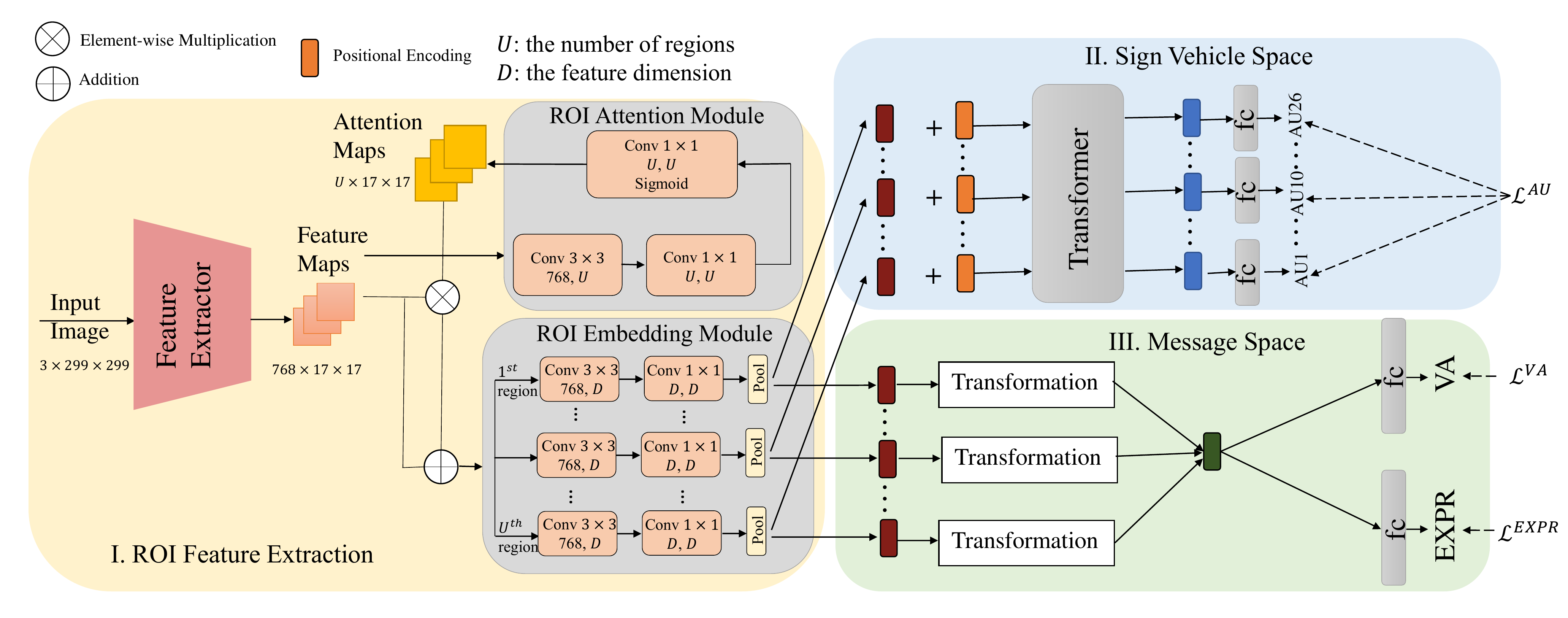}
    \caption{The architecture of our SMM-EmotionNet. }
    \label{fig:model}
\end{figure*}
Our main contributions are as follows:
\begin{itemize}
\setlength{\itemsep}{1pt}
\setlength{\parskip}{0pt}
\setlength{\parsep}{0pt}
    \item We propose a novel model architecture, the Sign-and-Message Multi-Emotion Net (SMM-EmotionNet) to learn the metric space of sign vehicles (\ie, the facial action units) and the metric space of the emotional messages (\ie, the facial expressions, valence and arousal).
    \item We employ the psychological prior knowledge into our architecture design, which showed superior performance than two baseline models. 
  
\end{itemize}

\section{Related Work}

Multitask emotion models, compared with uni-task emotion models, are more scarce, mainly because of the lack of datasets with the complete annotations of multiple emotion descriptors. Recently, there has been an increasing focus on predicting three emotion descriptors~\cite{kollias2021affect}, since the release of the Aff-wild and the Aff-wild2 dataset~\cite{kollias2019expression,kollias2019deep,zafeiriou2017aff}. 

In the ABAW2 Challenge~\cite{kollias2020analysing}, \cite{deng2021iterative} proposed a light-weight CNN-RNN model to predict three emotion descriptors in video sequences. In their uni-modal (visual modality) approach, they used a common architecture: a feature extractor shared by multiple tasks, followed by several branches. Each branch correspond to one task. Since they did not focus on the different properties of each task, they used similar branches for every one of them. Similar architecture was used in their approach~\cite{deng2020multitask} submitted to the first ABAW Challenge.

\cite{zhang2021prior} designed a dedicated architecture for three emotion descriptors prediction in the second ABAW Challenge. While using branches for three tasks, they also considered the relations between them. Therefore, they designed a serial manner of recognition process: AU $\rightarrow$ EXPR $\rightarrow$ VA. They claimed it as the recognition from local action units to global emotion states. Their work and our work both aim to employ the prior knowledge about the relations of multiple emotion descriptors into the architecture design. However, since we utilize the theory proposed by \cite{ekman1969repertoire}, the data flow of our model has two branches: one from the facial regions to the sign vehicles space (AU), another from the facial regions to the message space (EXPR and VA). 

\cite{kollias2019face} proposed an alternative "co-annotation" method utilizing prior knowledge about the relationship between facial expressions and action units, giving better robustness to data distribution shifts. They assigned annotations to missing labels based on the labels of other emotion descriptors. For example, "Happiness" is automatically assigned if AU12 (lip corner puller) and some other AUs are activated. They only considered relationships between discrete emotions and action units for co-annotation. Similar method was used in~\cite{kollias2021distribution}. Our approach is similar to the co-annotation method. However, we exploit prior knowledge differently. Rather than using prior knowledge to fill in missing annotations, we embed insights from psychological studies into the structure of our network. This captures correlations between different descriptors, but avoids the possible introduction of labelling errors based on the rule-based annotations assignment.  

\section{Methodology}

The architecture of our proposed Sign-and-Message EmotionNet is shown in Figure~\ref{fig:model}.

\subsection{Model Architecture}

\textbf{Sign Vehicle Space}. We define the sign vehicle spaces as the metric space of the facial action units. To learn representations for each AU, we use the Emotion Transformer proposed by~\cite{jacob2021facial}, who showed that exploiting intra-AU attention and inter-AU correlations are key components in AU prediction. The Emotion Transformer are shown in parts I and II of Figure~\ref{fig:model}. 

In part I, the feature extractor is an InceptionV3 model. The size of the extracted feature map is $17\times 17$ and it has 768 channels. The feature map is then fed into ROI attention modules to learn the spatial attention map for each region. The element-wise product between the attention map and the feature map is fed into the ROI embedding module to generate the feature vector of each region shown by dark red tensors in Figure~\ref{fig:model}. Both the ROI attention modules and embedding modules consist of multiple convolutional layers. More details about part I are in~\cite{jacob2021facial}. 

In part II, the features of each region are combined with positional encoding vectors and then fed into the the Transformer to learn correlations among the different regions corresponding to different AUs. Then, the output tensors shown by the blue color are the features on the sign vehicles space. This is also the metric space for each action unit. We simply use one fully connected (FC) layer to each AU metric space. The output of this FC layer is then applied with the Sigmoid function to predict the probability of the AU occurrence.

\textbf{Message Space}. The message space is the metric space for the facial expressions, valence and arousal. 
A psychological model, the Russell's circumplex model~\cite{russell1980circumplex}, supports our idea of learning the two emotion descriptors in a shared space. The Russell's circumplex model describes a 2-dimensional circular space, where the horizontal axis is valence and the vertical axis is arousal. The discrete facial expressions are mapped onto this 2D space, indicating their close relations with valence and arousal. Therefore, we design the message space shared by EXPR and VA. Given the features on the message space, we feed them into two FC layers to estimate EXPR and VA.

\textbf{Notations}. Given a facial image $x$, we first extract the regions of interest features. We denote the $u^{th}$ ROI feature vector as $f^{(u)}(x) \in \mathbb{R}^{D}$. $D$ is the feature dimension. For all $U$ ROI features, they are denoted as $F^{(U)}(x)  =\{f^{(u)}(x)\}_{u=1}^U$.

The transformer in part II (Figure~\ref{fig:model}) transforms the ROI features into the AU metric space. We denote the number of action units to be estimated as $H$. Note that when $U>H$, it means we have more facial regions than the number of action units to be estimated. We simply feed the first $H$ ROI features into the Transformer. The output of the Transformer is given by:

\begin{equation}
    S^{(H)}(x) = \Phi(F^{(H)}(x) + PE),
\end{equation}
where $PE$ denotes the positional encoding vector. $\Phi$ denotes the Transformer function. $S^{(H)}(x) = \{S^{(h)}(x)\}_{h=1}^H$, where $S^{(h)}(x)\in \mathbb{R}^D$ is the feature vector on the $h^{th}$ AU's metric space. We denote the weight matrix of the last FC layer for $h^{th}$ AU as $W_h\in\mathbb{R}^D$. The output of this FC layer is given by:

\begin{equation}
    y^{AU}_h   = \langle W_h, S^{(h)}(x) \rangle + b_h,
\end{equation}
where $\langle \cdot, \cdot \rangle$ is the inner product between the weight vector $W_h$ and the feature vector $S^{(h)}(x)$ . $b_h$ is the bias term.

For the message space learning in part III (Figure~\ref{fig:model}), the transformation module is learned for each region through the back-propagation. Since we do not have assumptions on the transformation module, we consider the simplest case: it is a linear transformation matrix. We denote the weight matrix as $A^{(u)}\in\mathbb{R}^{D\times D}$ for the $u^{th}$ region. The transformed vector on the message space can be denoted as:

\begin{equation}
    M^{(u)}(x) = \langle A^{(u)}, f^{(u)}(x) \rangle. 
\end{equation}

\begin{figure}
    \centering
    \includegraphics[width=0.7\linewidth]{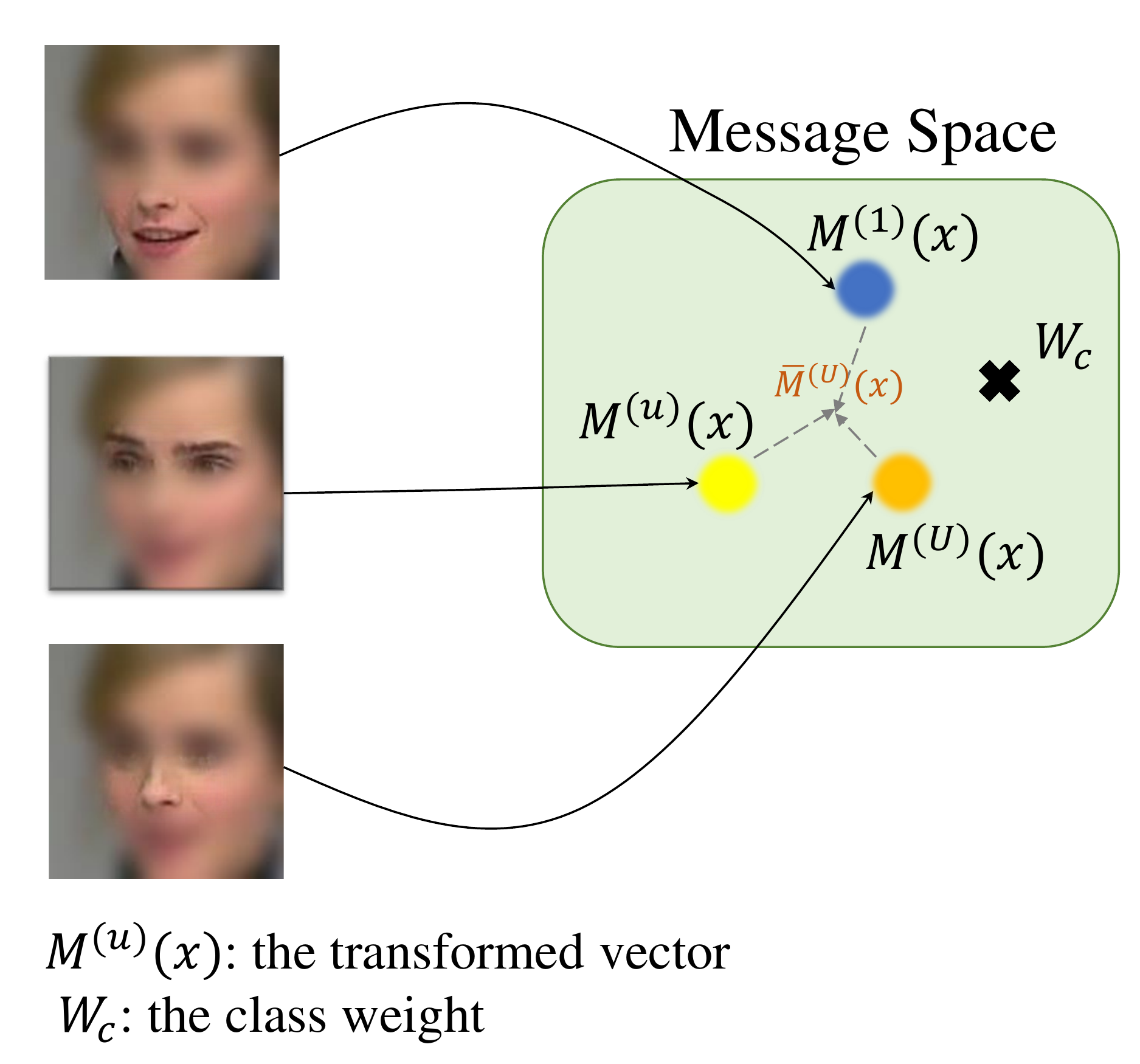}
    \caption{The process of averaging transformed vectors on the message space.}
    \label{fig:Message_Space}
\end{figure}

Given multiple transformed vectors, we can obtain the consensus by taking the average over all of them: $\bar{M}^{(U)}(x) = \frac{1}{U} \sum_u^U M^{(u)}(x)$. Next, we can compute the prediction of $c^{th}$ class: $\hat{y}_c = \langle W_c, \bar{M}^{(U)}(x)\rangle +b_c$, where $W_c$ is the weight vector for the $c^{th}$ class. Alternatively, we can take average over $\langle W_c, M^{(u)}(x)\rangle$ for all $u\leq U$. Because of the linearity of the last FC layer, the two approaches are equivalent. We use the first approach in this work. A diagram in Figure~\ref{fig:Message_Space} illustrates the process of feature projection and averaging on the message space.

\subsection{Losses}
 For the AU prediction task, the inference loss is the binary cross entropy loss. Given the input $x$, the AU prediction is denoted by $\hat{y}^{AU}$. The ground truth label for AU is denoted by $y^{AU}$. The inference loss of the AU task is given by:

\begin{equation} 
\begin{split}
   \mathcal{L}^{AU} (\hat{y}^{AU}, y^{AU}) = -&\frac{1}{U} \sum_i^U P^{AU}_i y^{AU}_i\log(\sigma(\hat{y}^{AU}_i)) +  \\
   &(1- y^{AU}_i)\log(1-\sigma(\hat{y}^{AU}_i)),
\end{split}
\end{equation}
where $P^{AU}_i$ is the weight of each AU for data balancing. It is computed from the training set data distribution. $P^{AU}_i$ equals to the number of negative samples divided by the number of positive samples for each AU. $\sigma(\cdot)$ represents the Sigmoid function.

For facial expression (EXPR) classification, we use a cross entropy loss as the inference loss shown in Equation~\ref{eq:Infer_loss_EXPR}. 
\begin{equation}
\begin{split}
   & \mathcal{L}^{EXPR} (\hat{y}^{EXPR}, y^{EXPR}) = \\
   & - \sum_i^C P^{EXPR}_i y^{EXPR}_i \log(\rho_i(\hat{y}^{EXPR})), 
   \label{eq:Infer_loss_EXPR}
\end{split}
\end{equation}
where $\rho_i(\hat{y}^{EXPR}) = \frac{exp(y^{EXPR}_i)}{\sum exp(y^{EXPR}_i)}$ denotes the Soft-max function. $ P^{EXPR}_i $ is the re-weighting factor, which is computed from the training set data distribution. 

Finally, for the valence and arousal (VA) prediction, we use the negative Concordance Correlation Coefficient (CCC) as the inference loss.
\begin{equation}
    \mathcal{L}^{VA}(\hat{y}^{VA}, y^{VA}) = 1- CCC^V + 1- CCC^A.
\end{equation}

To learn multiple tasks, we use a unweighted sum to combine different inference losses:

\begin{equation}
    L = \mathcal{L}^{AU} + \mathcal{L}^{EXPR} + \mathcal{L}^{VA}.
\end{equation}

\section{Experiments}

\subsection{Datasets}

The datasets provided by the ABAW3 challenge can be divided into two categories: the uni-task datasets and the multi-task learning (MTL) static datasets. 

The videos of the uni-task datasets are the same as the videos from the Aff-wild2 database~\cite{kollias2019expression}, including the AU subset, the EXPR subset and the VA subset. The challenge provides more annotations compared with the Aff-wild2 database. For example, in addition to seven facial expression categories, they annotated an addition facial expression, 'other' to the existing 7 facial expressions (six basic emotions plus neutral). In total, there are  548 videos of around 2.7 million frames in the uni-task datasets.

The MTL static dataset contains only a subset of frames from the Aff-wild2 dataset. Each of frame is annotated with 8 facial expressions, 12 action units, valence and arousal. In total, there are around $175,000$ images in the MTL static dataset.
 
In our experiments, we did not use the MTL static subset which has complete emotion annotations. We only used the video data from the three uni-task datasets. There are mainly two reasons. Firstly, our approach is proposed in order to alleviate the problem of lacking the complete annotations by regularizing the feature learning with prior knowledge. It is expected to have incomplete annotations as our training data. Secondly, we notice that the static MTL subset has overlapped videos with other three uni-task subsets in both the training and the validation set. It is problematic to use the three uni-task subsets and the MTL static dataset simultaneous because of the data leakage problem.

The AU subset we used contains around 1.8 million frames. The EXPR subset contains around 0.8 million frames. The VA subset contains around 1.8 million frames. We down-sampled these video frames by 8 times to reduce the training time.

\subsection{Training Details}

The aligned faces are provided by this Challenge. We resized the aligned faces to the size of $299\times299$ in pixels. 

The feature size of each ROI embedding is $D=16$. For the number of facial regions, we did a grid search: $U \in [12, 17, 27]$. $12$ is the number of AUs with annotations in the AU subset. $17$ is the number of AUs related to emotions found in \cite{morishima1993emotion}. $27$ is the number of major AUs in the facial action coding system~\cite{friesen1983emfacs}. Based on the grid-search results, we chose the number of regions as $U=17$. For those AUs without annotations, we did not compute the inference losses on them. The extra regions only provide certain degrees of freedom to the message space learning.
 
The optimizer we used is SGD. The momentum of SGD equals to $0.9$. The initial learning rate is $10^{-3}$. A cosine annealing learning rate schedule is used to improve convergence. The total number of training iterations is $3\times10^5$.

\subsection{Evaluation Metric}
For AU detection task, we use the averaged F1 score of 12 AUs to evaluate its performance. For discrete emotion classification, we use the averaged macro F1 score of eight facial expressions (six basic emotions, neutral and others) as the evaluation metric. For continuous emotion prediction, we use CCC as the evaluation metric. 

\section{Results}

\subsection{Static Approach}

\textbf{Comparison with baseline models}

In the ABAW3 challenge, the official baseline model provided by the challenge organizer is a single-task CNN model (ResNet50 model for VA and VGG16 model for EXPR and AU)~\cite{kollias2021analysing}. They changed the size of the final FC layer according to the task. We also compared our SMM-EmotionNet with a multi-task baseline model: an InceptionV3 feature extractor followed by parallel branches of FC layers corresponding to different tasks. In each branch, the weight size of consecutive FC layers are: $768\times 16 \rightarrow 16 \times C$. $C$ is the number of classes in the task prediction. 

Compared with both single-task and multitask baseline models, our model showed superior performance on every task.

\begin{table}[t]
    \centering
    \begin{tabular}{c|c|c|c|c}
    \hline
        Model & F1-AU & F1-EXPR&  CCC-V & CCC-A\\
       \hline
       \hline
        VGG-Face& 0.39&  0.23&  0.31& 0.17 \\
         \hline
         InceptionV3& 0.536 & 0.489 & 0.441 & 0.485\\
         \hline
         \hline
        Ours & \textbf{0.548 }& \textbf{0.518} & \textbf{0.447} & \textbf{0.499}\\
        \hline
    \end{tabular}
    \caption{The comparison between our static model and two baseline models (also static). The results are evaluated on validation sets of the AU, EXPR and VA subset from the Aff-wild2 dataset. F1-AU stands for the unweighted F1 score of 12 action units. F1-EXPR stands for the macro F1 score of eight facial expressions. CCC-V stands for the CCC value of valence. CCC-A stands for the CCC value of arousal.}
    \label{tab:baseline_vs_ours}
\end{table}

\subsection{Temporal Approach}
We considered a simple temporal smoothing method for the features on the message space or the sign vehicles space in order to filter high-frequency noises. For the feature vector $\epsilon_t =M^{(u)}(x_t)$ or $\epsilon_t =S^{(u)}(x_t)$ at the time step $t$, we smoothed it with the this function:

\begin{equation}
    \Lambda_t = \frac{1}{1+\mu} (\epsilon_t  + \mu \Lambda_{t-1}).
\end{equation}

$ \Lambda_t$ is the feature vector at the time step $t$ after smoothing. $\mu$ is a hyper-parameter which determines the smoothness. We fed the $ \Lambda_t$ into the three FC layers for classification or regression without re-training our model.

To find the best $\mu$ through grid search, we performed three cross-validation on the validation sets. The search region is $\mu = \{0, 1, 2, 3, 4, 5, 6, 7, 8, 9, 10\}$. Given the cross validation results, we chose the $\mu$ for AU task to be $7$, while the $\mu$ for EXPR and VA to be $9$. 

The three-fold validation set performance is shown in :

\begin{table}[]
    \centering
    \begin{tabular}{c|c|c|c}
    \hline
       Fold ID& F1-AU & F1-EXPR & CCC-VA (Average)\\
        \hline
        \hline
        1 & 0.463 & 0.428 & 0.542\\
        2 & 0.586 & 0.552 & 0.577\\
        3 & 0.576 & 0.450 &  0.558\\
        \hline
    \end{tabular}
    \caption{The three-fold cross validation results given the optimal $\mu$ on each of the validation sets in the AU, EXPR and VA subset. CCC-VA stands for averaged CCC values of valence and arousal.}
    \label{tab:temporal_model}
\end{table}

\subsection{Multi-task Learning Performance}

As we proposed a multitask solution, the challenge track we participated in is the MTL (Multitask Learning) challenge track. The performance metric in the MTL challenge is defined as:

\begin{align}
    P  = &\frac{1}{2}(CCC^{V} + CCC^{A}) + \frac{1}{8}\sum_i^8 F_1^{EXPR_i}  + \nonumber\\
    &\frac{1}{12}\sum_j^{12} F_1^{AU_j}
    \label{eq:mtl_metric}
\end{align}

\begin{table}[]
    \centering
    \begin{tabular}{c|c|c}
       Model  & Val sets (uni-task) & Test set (MTL) \\
       \hline
       ours (static)  & 1.539 & 1.104\\
       ours (temporal) & 1.664& 1.113\\
       \hline
    \end{tabular}
    \caption{The multitask evaluation metric (Equation~\ref{eq:mtl_metric}) on the validation sets of uni-task subsets and the test set of the MTL static dataset.}
    \label{tab:mtl_metric_table}
\end{table}

Using the Equation~\ref{eq:mtl_metric} as the evaluation metric, we evaluated the performance of our static and temporal approaches on the validation sets of the three uni-task datasets and the test set of the static MTL dataset. The results are shown in Table~\ref{tab:mtl_metric_table}. The temporal smoothing method boosted the overall multitask performance slightly compared with the static approach. We think there might be better methods to learn the temporal relations than the simple temporal smoothing.

\section{Conclusion}
In this paper, we propose a novel architecture for multiple emotion descriptor recognition and feature learning. We designed distinct architectures for the feature space of sign vehicles (\eg, facial action units) and the feature space of emotional messages (\eg, facial expressions, valence and arousal) according to their properties. Our code implementation is available at \url{https://github.com/wtomin/ABAW3_MultiEmotionNet}. 

\bibliographystyle{named}
\bibliography{ijcai22}

\end{document}